\documentclass{article}

\usepackage{graphicx}
\usepackage{amsmath}

\usepackage[preprint]{neurips_2025}

\usepackage[utf8]{inputenc} 
\usepackage[T1]{fontenc}    
\usepackage{hyperref}       
\usepackage{url}            
\usepackage{booktabs}       
\usepackage{amsfonts}       
\usepackage{nicefrac}       
\usepackage{microtype}      
\usepackage{xcolor}         

\title{Game-invariant Features Through Contrastive and Domain-adversarial Learning}

\author{%
  Dylan Kline  \\
  Hajim School of Engineering \& Applied Sciences: Department of Computer Science \\
  University of Rochester\\
  \texttt{dkline4@u.rochester.edu} \\
}

\begin{document}

\maketitle

\begin{abstract}
Foundational game-image encoders often overfit to game-specific visual styles, undermining performance on downstream tasks when applied to new games. We present a method that combines contrastive learning and domain-adversarial training to learn game-invariant visual features. By simultaneously encouraging similar content to cluster and discouraging game-specific cues via an adversarial domain classifier, our approach produces embeddings that generalize across diverse games. Experiments on the Bingsu game-image dataset (10,000 screenshots from 10 games) demonstrate that after only a few training epochs, our model’s features no longer cluster by game, indicating successful invariance and potential for improved cross-game transfer (e.g. glitch detection) with minimal fine-tuning. This capability paves the way for more generalizable game vision models that require little to no retraining on new games.
\end{abstract}

\section{Introduction}
\label{sec:intro}

Modern foundation models for game images (e.g. large ConvNets or vision transformers pre-trained on broad data) tend to group features by the game’s visual style\cite{trivedi2021Cont}. In other words, images from the same game (or genre) cluster together in the feature space due to superficial stylistic similarities, rather than semantic content. This behavior is problematic for downstream game-specific tasks such as glitch detection – a model might fail to recognize similar glitches in a new game if its features are overly tuned to the original game’s art style. Indeed, neural models trained on raw game pixels often capture visual style differences rather than the underlying game content \cite{trivedi2021Cont}, leading to poor generalization even across games of the same genre \cite{trivedi2021Cont}. Ideally, we want representations that capture game-agnostic concepts so that we can generalize to any new game with very little training data or effort. For instance, a vision model should recognize a graphical glitch or a common object type across different games, irrespective of whether one game is a 2D pixel-art platformer and another is a 3D realistic shooter. Achieving such generalization across unseen games is a key challenge for game AI and analytics, as it would enable reusing learned models for new titles without extensive retraining.

In this work, we tackle the above problem by learning game-invariant features, feature representations that are informative for high-level visual understanding but do not carry information specific to any particular game’s identity. We propose to accomplish this through a combination of contrastive learning and domain-adversarial training. Contrastive learning has shown promise in forcing representations to capture meaningful content by bringing similar images together and pushing dissimilar ones apart in the feature space. However, by itself, contrastive learning on game images might still allow the model to inadvertently use game-specific cues, since many images from the same game will naturally share content. To explicitly remove game-specific information, we introduce a domain-adversarial objective, we train a game (domain) classifier on the learned features and simultaneously train the feature encoder to “trick” this classifier. In doing so, the encoder learns to discard those aspects of the image that reveal which game it came from, while preserving the broader features that can contribute to semantic understanding across games. Our goal is a versatile visual encoder that can be applied to previously unseen games with minimal, if any, additional training, thus moving closer to “universal” game vision models.






\section{Related Work}
\label{sec:background}

\subsection{Game Representation Learning}

Trivedi et al. \cite{trivedi2021Cont} introduced a contrastive learning approach to obtain generalized representations of video games. Their work, Contrastive Learning of Generalized Game Representations, trained encoders on over 100k images from 175 games spanning 10 genres. The key finding was that contrastive training caused the learned features to ignore visual stylistic details and focus on game content, yielding a more meaningful separation of games by genre/content rather than surface appearance. In other words, the contrastively learned embedding grouped games in a way that reflected gameplay or genre similarity, as opposed to a standard supervised model which clustered games by art style. This result suggests that contrastive objectives can help learn content-centric features in games, improving generalization within and across game genres. However, Trivedi et al.’s method did not explicitly remove all game-specific information – some residual clustering by game or genre could remain, since no adversarial mechanism was used to enforce invariance. Our work builds on this idea by adding an adversarial component to push the encoder towards true game-agnostic representations.

\subsection{Domain-Adversarial Learning}

Domain-adversarial training is a technique commonly used in unsupervised domain adaptation to learn features that are invariant to the domain (or dataset) of origin. Notably, Huang \cite{huang2023time} explored a contrastive adversarial domain disentangled network for time-series data. In that work, dubbed CADDN, the model jointly learns domain-invariant and domain-specific features by employing two complementary objectives. A domain classifier attempts to distinguish the source vs. target time-series domain, while the feature extractor is trained adversarially (via a gradient reversal layer) to fool the domain classifier, thus producing domain-invariant features. Simultaneously, a contrastive loss encourages better separation of the domain-specific features, improving stability. Huang’s results showed that adding a contrastive loss on top of adversarial domain training improved adaptation performance, as it retained useful information that pure invariant-feature learning might ignore. This idea of combining contrastive and adversarial objectives informs our approach. We similarly aim to disentangle what is shared vs. specific across domains (games), except in our case the “domains” are different games and the data are images rather than sensor time-series. By drawing inspiration from both Trivedi et al.’s \cite{trivedi2021Cont} contrastive game representation learning and Huang’s \cite{huang2023time} adversarial contrastive framework, we design a model that uncovers invariant features across games while preserving meaningful variability.

\section{Methods}
The core idea of our method is to use a hybrid objective that blends contrastive learning with domain-adversarial training to learn features that contain useful visual information but are independent of the game identity. Formally, we train an encoder network $E(\cdot)$ such that its image embeddings are (1) effective at distinguishing different images based on content (via a contrastive loss), and (2) unable to be reliably classified by game (via an adversarial domain loss). The latter forces the encoder to “trick” a game classifier – any features that would reveal the source game are suppressed, leaving only game-invariant cues. By balancing these two forces, the encoder can retain important semantic features (e.g. objects, scene layout) that occur across games, while discarding style or texture features unique to each game.

This combination is non-trivial, if we only did adversarial training, the encoder might collapse to some trivial representation that loses too much information (to confuse the domain classifier); the contrastive term prevents that by actively encouraging informative representations. Conversely, if we only did contrastive learning, the encoder might still cluster images by game (since many images from the same game share visual attributes), so the adversarial term explicitly breaks those clusters. Our approach, therefore, finds a sweet spot where the learned embedding space groups images by higher-level similarity rather than by game/style. In practice, we implement the adversarial objective using a Gradient Reversal Layer (GRL) as in domain-adversarial networks: during backpropagation, the gradient from the domain-classifier is multiplied by -1 before passing into the encoder, effectively making the encoder’s objective the inverse of the classifier’s.

\subsection{Dataset}

We train and evaluate our method on the Bingsu Gameplay Images dataset \cite{bingsu2022gameplayimages}. This dataset (hosted on HuggingFace) consists of 10,000 screenshots collected from gameplay videos of 10 popular and stylistically varied video games: Among Us, Apex Legends, Fortnite, Forza Horizon, Free Fire, Genshin Impact, God of War, Minecraft, Roblox, and Terraria. Each game contributes 1,000 screenshot images, with a fixed resolution of 640×360 pixels. 

The games in the dataset span a wide range of genres and art styles – from cartoonish 2D and pixel art (e.g. Terraria, Among Us) to realistic 3D (e.g. Forza, God of War) and stylized 3D (Fortnite, Genshin). This diversity is ideal for our purposes, an encoder that truly captures game-agnostic features should be able to find commonalities between, say, a blocky voxel image from Minecraft and a cel-shaded anime-style image from Genshin, if they share some structural or semantic similarity, rather than grouping images purely by the game label. We use the game title labels in this dataset solely for defining the domain (game) classification task in our adversarial training. Importantly, we do not use any task-specific labels (e.g., glitch vs. normal) during representation learning – the training is self-supervised contrastive, with the only supervised signal being the game domain which is used adversarially. Thus, our learned encoder is not biased toward any particular downstream task and can be evaluated generally.

\section{Experiments}

\subsection{Training and Architecture}
\begin{figure}[h]
  \centering
  \includegraphics[width=0.7\linewidth]{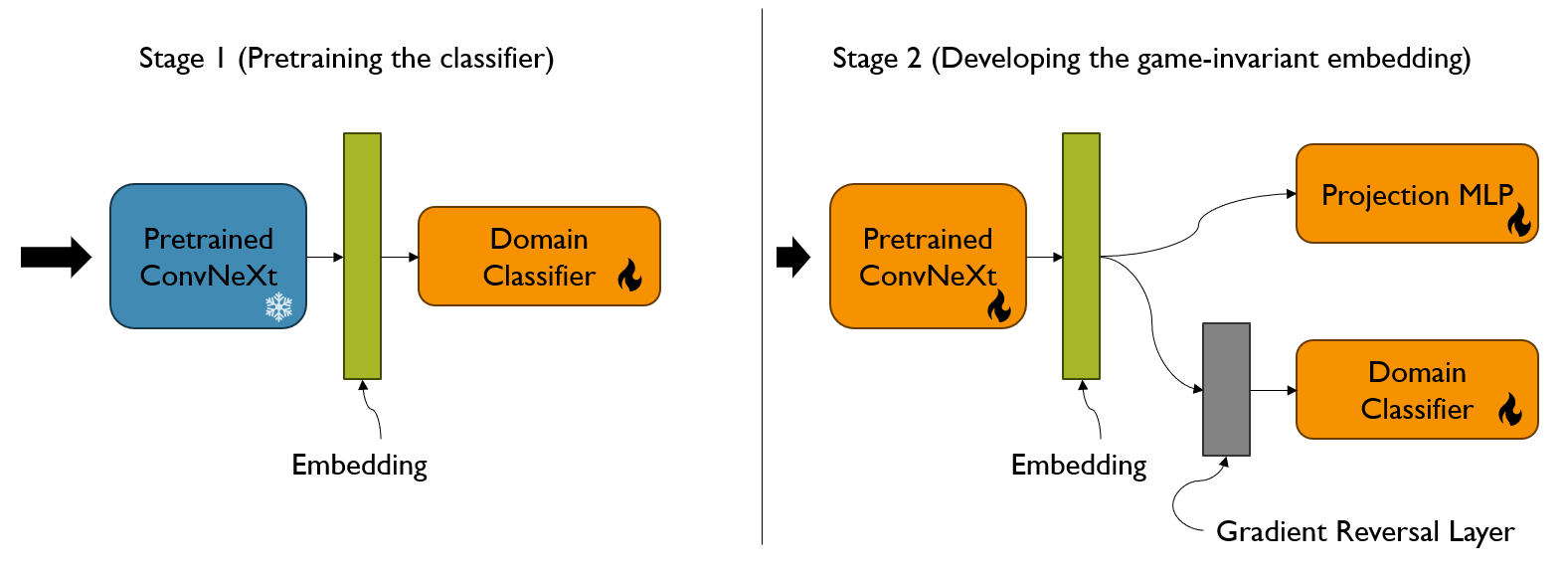}
  \caption{Two-stage training for game-invariant features. Stage 1 (left): Frozen ConvNeXt (blue) extracts embeddings fed to a trainable domain classifier (orange) to verify game-specific information. Stage 2 (right): Fine-tune ConvNeXt and a projection MLP head (orange) with a GRL-attached domain classifier (gray). Jointly optimize a contrastive loss on MLP outputs and an adversarial loss via the GRL to suppress game cues and produce game-invariant representations.}
  \label{fig:retention_8}
\end{figure}

We adopt a two-stage training process to learn our game-invariant encoder, as illustrated in Figure 1. The stages are as follows:

\subsubsection{Stage 1 – Pretraining the Game Classifier}
First, we assess how much game-specific information is present in off-the-shelf features. We take a ConvNeXt network pre-trained on ImageNet as a fixed feature extractor. We feed all 10k images through this frozen ConvNeXt and obtain embeddings (e.g. from a late pooling layer). We then train a lightweight domain classifier on top of these embeddings to predict which game each image came from. This is a standard 10-way classification problem (one class per game). 

We found that the domain classifier achieves a high accuracy on the training and validation sets, indicating that the ConvNeXt’s native features are highly discriminative of game identity. In other words, the pre-trained encoder naturally groups images by visual style/texture, which is not surprising, ImageNet features tend to pick up on color palettes, textures, etc., that often correlate with each game’s graphics style. 

This result reinforces the premise that without intervention, foundation model features are entangled with game-specific style, so we need to actively mitigate that. 

\subsubsection{Stage 2 – Learning a Game‐Invariant Embedding}

In stage 2, we fine-tune the ConvNeXt encoder with two complementary objectives to strip out game cues while retaining visual content. We attach a small projection MLP on top of the encoder to produce the final embedding used for contrastive learning (following common practice in self-supervised learning to have a projection head)\cite{trivedi2021Cont}. We reuse a domain classifier of the same architecture as in Stage 1 to provide the adversarial signal. Crucially, we insert a Gradient Reversal Layer (GRL) between the encoder and the domain classifier. 

During forward pass, the GRL acts as an identity function (features pass through unchanged), but during backpropagation it multiplies the gradients by -1 before flowing into the ConvNeXt and projection MLP. This implements the adversarial objective, the domain classifier is trained to minimize classification loss, whereas the encoder and projection head combination receives the negated gradient, meaning it is effectively trained to maximize the domain classification loss, i.e. to make the features as uninformative about the game as possible. The adversarial objective is defined as follows:

\begin{equation}
\mathcal{L}_{\mathrm{dom}}
= \mathbb{E}_{(x,d)}\Bigl[-\sum_{g=1}^{G}
     \mathbf{1}_{[d=g]}\,\log D_g\bigl(\mathrm{GRL}(z(x))\bigr)\Bigr].
\label{eq:adv_loss}
\end{equation}

At the same time, we apply a contrastive learning objective on the encoder. Specifically, we adopt a SimCLR-style contrastive loss, where each input image is heavily augmented twice (random crops, color jitter, etc.) to create two “views,” and we encourage the encoder to map these two views to nearby points in the embedding space while pushing apart embeddings of different images \cite{trivedi2021Cont}. We use the NTXent contrastive loss for this purpose \cite{sohn2016improved}, which is defined as follows for one positive pair (i, j):

    \begin{equation}
\ell_{i,j}^{\mathrm{con}}
= -\log
  \frac{\exp\!\bigl(\mathrm{sim}(z_i, z_j)/\tau\bigr)}
       {\displaystyle\sum_{\substack{k=1 \\ k\neq i}}^{2N}
         \exp\!\bigl(\mathrm{sim}(z_i, z_k)/\tau\bigr)}
\label{eq:per_pair_contrastive}
\end{equation}

The contrastive loss ensures that the encoder captures high-level similarities (images of the same scene or object should come together) and avoids collapse, while the adversarial loss ensures that none of these similarities are simply “both are from Game X.” We train Stage 2 for a small number of epochs and monitor the domain classifier’s accuracy. As training progresses, the domain classifier’s performance should drop towards chance level \(10\%\) for 10 classes, signifying that the features no longer carry meaningful game-identifying information. Meanwhile, the contrastive loss steadily decreases, indicating the features are learning structure and commonalities across the diverse images. Thus, the total objective function is,

\begin{equation}
\mathcal{L}_{\text{total}}
= \mathcal{L}_{\mathrm{con}}
+ \lambda\,\mathcal{L}_{\mathrm{dom}}
\label{eq:total_loss}
\end{equation}

The outcome of Stage 2 is a fine-tuned ConvNeXt that produces game-invariant yet descriptive image embeddings. These embeddings can then be used for any downstream task. We expect that models built on top of this embedding (for example, a glitch detector) will be far more generalizable to new games, since the input features are not biased to any specific game.

\section{Results}

\begin{figure}[h]
  \centering
  \includegraphics[width=0.6\linewidth]{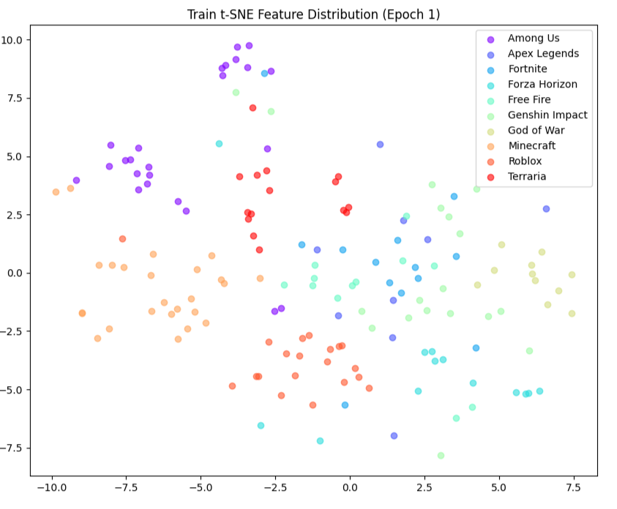}
  \caption{t-SNE visualization of the initial ConvNeXt feature embeddings (Epoch 1), colored by game title.}
  \label{fig:games_not}
\end{figure}

As shown in Figure~\ref{fig:games_not}, before applying our method the embedding space is clearly segregated by game. The scatter plot was generated using features from the frozen ConvNeXt encoder prior to any contrastive or adversarial training. The tight clustering of same-colored points demonstrates that the model’s representation is picking up on game-specific characteristics. For instance, Minecraft images (light orange) cluster separately from Fortnite images (cyan), presumably because of obvious differences in pixelation and color tone, not necessarily because they contain fundamentally different content. This confirms that without intervention, a model would likely overfit to the training games – any new game’s images would be embedded arbitrarily, and features useful for one game might not transfer to another. In terms of downstream impact, a glitch detection model built on such features might only recognize glitches in the context of the specific game it learned, since the notion of “what’s normal vs. glitch” could be entangled with the game’s look.

\begin{figure}[h]
  \centering
  \includegraphics[width=0.6\linewidth]{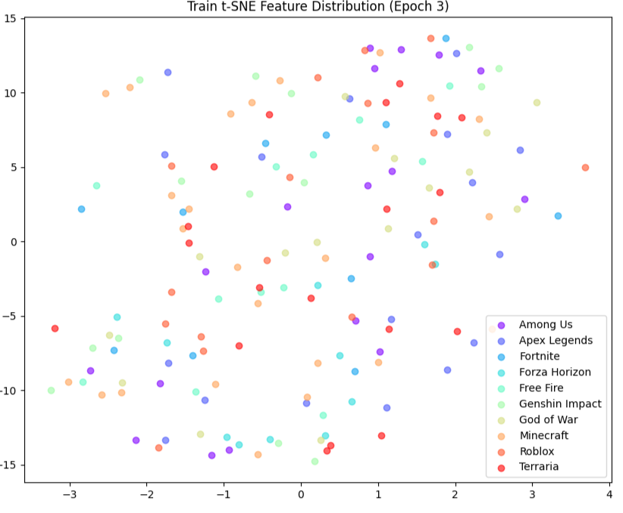}
  \caption{t-SNE visualization of the learned game-invariant embeddings at Epoch 3 after our contrastive + adversarial training. Points are again colored by game, but now the clusters have largely dispersed and different colors are intermixed.}
  \label{fig:games_in}
\end{figure}

After applying our two-stage training approach, the feature space is significantly reorganized as seen in Figure~\ref{fig:games_in}. We observe that game-specific clustering has been largely eliminated. The points form a more continuous cloud where colors (games) are mixed together. Crucially, this does not mean the features are meaningless – rather, it means that images are now grouped by higher-level visual patterns or content that can appear in multiple games. 

For instance, a screenshot of a bright outdoor scene in Fortnite might end up near a bright outdoor scene in Forza Horizon, even though one is a shooter and the other a racing game, because the encoder has learned to pay attention to scene-level features instead of game identity. Likewise, a 2D pixel-art character in Terraria might embed near a 2D cartoon character from Among Us if they share shape or color properties, whereas previously all Terraria images would cluster together solely due to their common pixel style. Quantitatively, the domain classifier’s accuracy on these embeddings drops to roughly random (around \(10–15\%\), from nearly \(100\%\) initially), confirming that the game label is no longer predictable – an explicit sign of domain invariance.

\section{Conclusions}
Our results show that contrastive and domain-adversarial training successfully produce visual features invariant to specific games, requiring only minimal additional training. This approach enables robust generalization across diverse games, highlighting its potential as a foundational method for universal game vision models. 

For future work, we plan to rigorously evaluate our game-invariant encoder on downstream tasks such as glitch detection and game playing to further confirm its utility across unseen games. Additionally, we aim to explore hybrid models combining our invariant features with targeted genre-specific embeddings as a way to efficiently adapt quality assurance systems to new unseen games.

\section*{Author Contributions}
D.K. contributed to the research as the sole researcher behind the experiments, methods, creation of the manuscript, code, and experimental analysis.

{
\small
\bibliographystyle{plainnat}
\bibliography{library}
}





\end{document}